\documentclass[letterpaper]{article} 
\usepackage{aaai23}  
\usepackage{times}  
\usepackage{helvet}  
\usepackage{courier}  
\usepackage[hyphens]{url}  
\usepackage{graphicx} 
\usepackage{amsmath}
\usepackage{amssymb}

\usepackage{natbib}  
\usepackage{caption} 
\usepackage{algorithm}
\usepackage{algorithmic}
\usepackage{amsmath,amssymb,stmaryrd}
\usepackage{arydshln}
\usepackage{newfloat}
\usepackage{listings}
\usepackage{amsfonts}
\DeclareCaptionStyle{ruled}{labelfont=normalfont,labelsep=colon,strut=off} 
\floatstyle{ruled}
\newfloat{listing}{tb}{lst}{}
\floatname{listing}{Listing}
\usepackage[switch]{lineno}
\usepackage{xspace}
\usepackage{colortbl}

\usepackage{booktabs}  
\usepackage{threeparttable}  
\usepackage{multicol}  
\usepackage{multirow}  
\usepackage{mathrsfs}
\usepackage{stfloats}

\newcommand{\graphix}{\textsc{Graphix}\xspace}

\newcommand{\spider}{\textsc{Spider}\xspace}

\newcommand{\syn}{\textsc{Syn}\xspace}
\newcommand{\real}{\textsc{Realistic}\xspace}
\newcommand{\dk}{\textsc{Dk}\xspace}

\newcommand{\picard}{\textsc{Picard}\xspace}

\def\etal{\emph{et al.}}
\def\eg{\emph{e.g.}}
\def\ie{\emph{i.e. }}

\title{Graphix-T5: Mixing Pre-Trained Transformers with \\ Graph-Aware Layers for Text-to-SQL Parsing}
\author{
    Jinyang Li\textsuperscript{\rm 1,2}\thanks{\quad Work done during an intern at Alibaba DAMO Academy.
    },
    Binyuan Hui\textsuperscript{\rm 2},
    Reynold Cheng\textsuperscript{\rm 1,5}\thanks{\quad \small Corresponding authors are Reynold Cheng and Yongbin Li.},
    Bowen Qin\textsuperscript{\rm 3},
    Chenhao Ma\textsuperscript{\rm 4},
    Nan Huo\textsuperscript{\rm 1}, \\
    Fei Huang\textsuperscript{\rm 2}, 
    Wenyu Du\textsuperscript{\rm 1},
    Luo Si\textsuperscript{\rm 2},
    Yongbin Li\textsuperscript{\rm 2} \footnotemark[2]
}

\affiliations{
    \textsuperscript{\rm 1}The University of Hong Kong 
    \textsuperscript{\rm 2}DAMO Academy, Alibaba Group \\
    \textsuperscript{\rm 3}Shenzhen Institute of Advanced Technology, Chinese Academy of Sciences \\
    \textsuperscript{\rm 4}The Chinese University of Hong Kong (Shenzhen)
    \textsuperscript{\rm 5}Guangdong–Hong Kong-Macau Joint Laboratory\\
    \small \texttt{\{jl0725,huonan,wenyudu\}@connect.hku.hk, ckcheng@cs.hku.hk,} \\ 
\small \quad \texttt{bw.qin@siat.ac.cn, machenhao@cuhk.edu.cn,}\\
\small \quad \texttt{\{binyuan.hby,f.huang,luo.si,shuide.lyb\}@alibaba-inc.com}
}

\usepackage{bibentry}

\begin{document}
\maketitle

\begin{abstract}
The task of text-to-SQL parsing, which aims at converting natural language questions into executable SQL queries, has garnered increasing attention in recent years, as it can assist end users in efficiently extracting vital information from databases without the need for technical background. 
One of the major challenges in text-to-SQL parsing is domain generalization, \ie, how to generalize well to unseen databases. 
Recently, the pretrained text-to-text transformer model, namely T5, though not specialized for text-to-SQL parsing, has achieved state-of-the-art performance on standard benchmarks targeting domain generalization.
In this work, we explore ways to further augment the pre-trained T5 model with specialized components for text-to-SQL parsing.
Such components are expected to introduce structural inductive bias into text-to-SQL parsers thus improving model's capacity on (potentially multi-hop) reasoning, which is critical for generating structure-rich SQLs. 
To this end, we propose a new architecture \textbf{\graphix-T5}, a mixed model with the standard pre-trained transformer model augmented by some specially-designed graph-aware layers.
Extensive experiments and analysis demonstrate the effectiveness of \graphix-T5 across four text-to-SQL benchmarks: \textsc{\textbf{Spider}}, \textsc{\textbf{Syn}}, \textsc{\textbf{Realistic}} and \textsc{\textbf{Dk}}.
\graphix-T5 surpass all other T5-based parsers with a significant margin, achieving new state-of-the-art performance. 
Notably, \graphix-T5-large reach performance superior to the original T5-large by 5.7\% on exact match (EM) accuracy and 6.6\% on execution accuracy (EX). This even outperforms the T5-3B by 1.2\% on EM and 1.5\% on EX. 
\end{abstract}

\section{Introduction}
Relational database, serving as an important resource for users to make decision in many fields, such as health care, sports, and entertainment, has emerged frequently because of the big data era.
It is efficient for data users to access the information from databases via structured query language, \eg, SQL. Despite its effectiveness and efficiency, the complex nature of SQLs leads to extremely expensive learning efforts for non-technical users.
Therefore, text-to-SQL~\citep{ijcai2018-553, geoquery, xu2017sqlnet, yu-etal-2018-typesql, yaghmazadeh2017sqlizer}, aiming to convert natural language instructions or questions into SQL queries, has attracted remarkable attention. 

\begin{figure}
    \centering
    \includegraphics[width=0.45\textwidth]{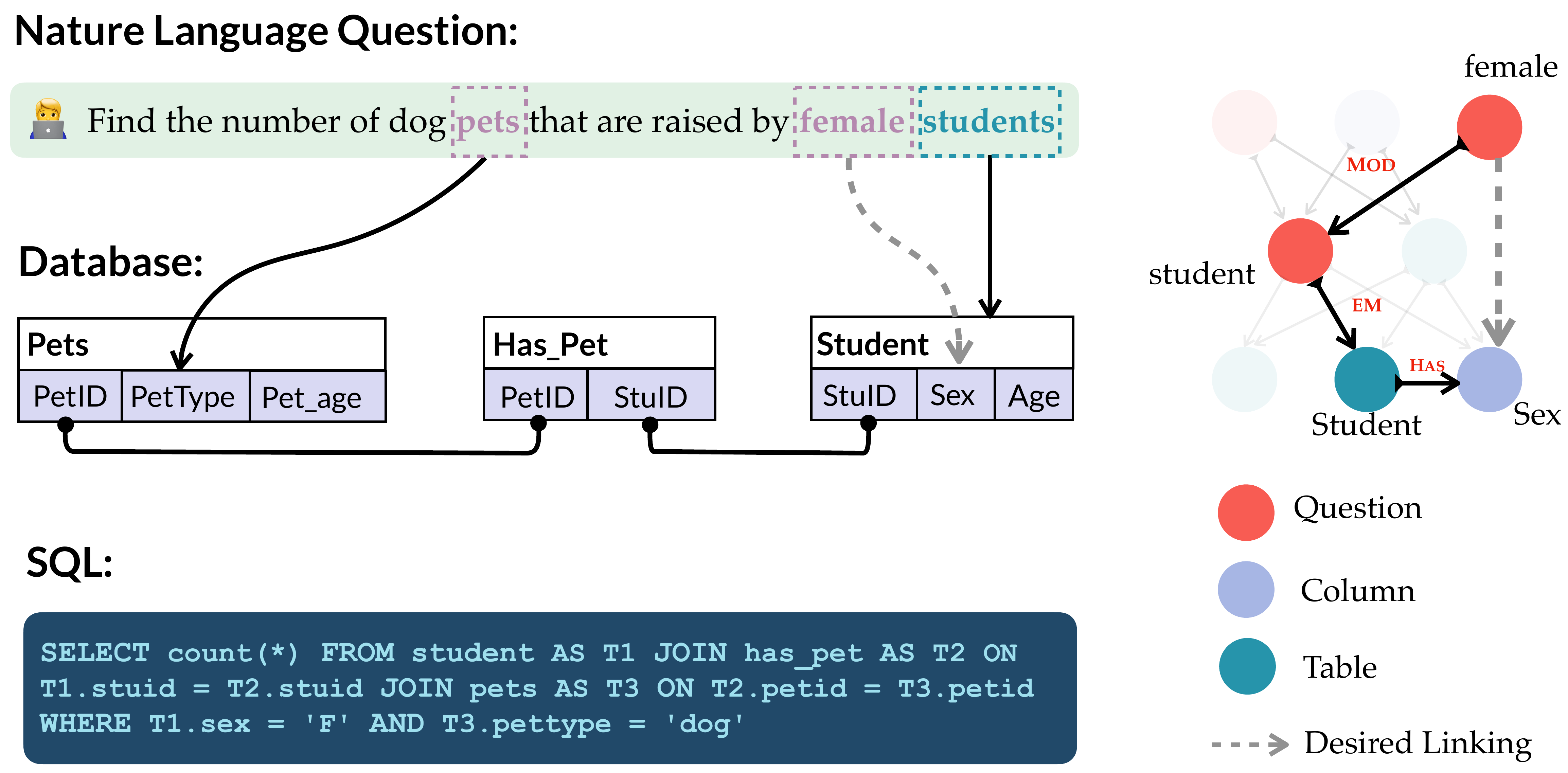}
    \caption{This is an an illustration of cross-domain text-to-SQL challenge. The link between the target column \texttt{sex} and the token \texttt{female} is highly desired but extremely challenging for the model to capture, especially when domain-specific data or effective rules is absent. However, this dilemma can be mitigated by a multi-hop reasoning path (\texttt{female} $\stackrel{\textsc{Mod}} \longrightarrow$ \texttt{student} $\stackrel{\textsc{Em}}\longrightarrow$ \texttt{Student} $\stackrel{\textsc{HAS}}\longrightarrow$ \texttt{Sex}).}
    \label{intro1}
    \vspace{-0.5cm}
\end{figure}

\begin{figure*}
    \centering
    \includegraphics[width=0.8\textwidth]{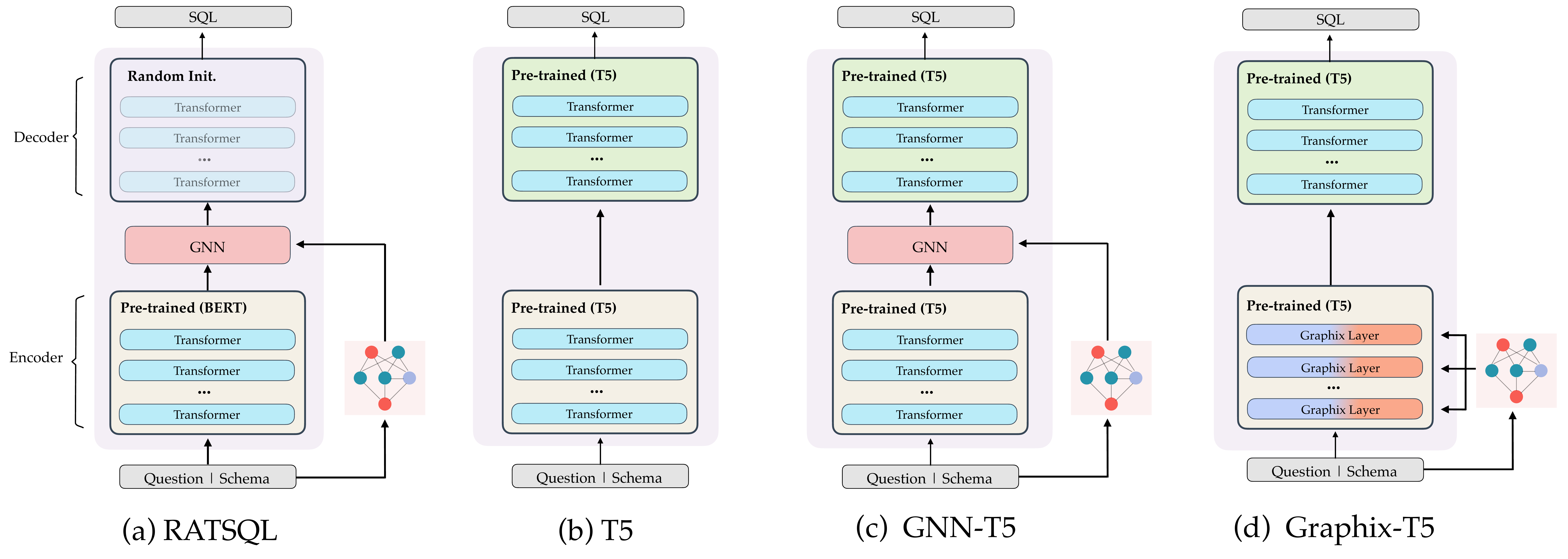}
    \caption{Graphical illustration of existing methods \textbf{(a) RATSQL} {[pre-trained BERT-encoder $\rightarrow$ graph-based module $\rightarrow$ randomly initialized decoder]}. \textbf{(b) T5} {[pre-trained T5-encoder $\rightarrow$ pre-trained T5-decoder]} and the proposed variant \textbf{(c) GNN-T5} {[pre-trained T5-encoder $\rightarrow$  graph-based module $\rightarrow$ pre-trained T5-decoder]} \textbf{(d) \graphix-T5} {[semi-pre-trained graphix-module $\rightarrow$ pre-trained T5-decoder]}.}
    \label{t5}
    \vspace{-0.3cm}
\end{figure*}

In this work, we explore the challenging cross-domain setting where a text-to-SQL parser needs to achieve \textit{domain generalization}, \ie, the ability to generalize to domains that are unseen during training. 
Achieving this goal would, in principle, contribute to a universal natural language interface that allows users to interact with data in arbitrary domains. The major challenge towards domain generalization \citep{wang-etal-2020-rat, lgesql,proton,sadga, s2sql} is that generating structure-rich SQLs requires (potentially multi-hop) \textbf{reasoning}, \ie the ability to properly contextualize a user question against a given database by considering many explicit relations (\eg, table-column relations specified by database schema) and implicit relations (\eg, whether a phrase refers to a column or table). 
Figure \ref{intro1} shows an introductory example of multi-hop reasoning in the text-to-SQL parsing and Figure \ref{case} presents two more detailed cases.

From the modeling perspective, there are two critical dimensions along which we can differentiate current text-to-SQL parsers. The first is how to effectively imbue relational structures (both explicit and implicit) in the form of graphs into neural networks, and the second is how to take the most advantage of pre-trained models (\eg T5 \citep{t5-jmir-2020}). These two dimensions are inter-connected and form a spectrum of methods. On one end of the spectrum, PICARD \citep{PICARD} uses the original pre-trained T5 model by linearizing database schemas into sequences, hoping that T5 can successfully capture the underlying relational structures. On the other end of the spectrum, RAT-SQL \citep{wang-etal-2020-rat} only utilizes pre-trained encoders (\eg, BERT \citep{devlin-etal-2019-bert})
and explicitly captures desired relations via specialized relation-aware models. However, more powerful encoder-decoder based pre-trained models are not exploited in this framework, but relational structures are accommodated at most. In this work, we explore the cross zone where the encoder-decoder based pre-trained models (specifically T5) and relation-aware encodings are deeply coupled in favor of better domain generalization.
We first observe that naively adding a relational graph-based module in the middle of T5, resulting in a `T5-encoder $\rightarrow$ graph-based module $\rightarrow$ T5-decoder architecture' (see also Figure~\ref{t5}(c), namely GNN-T5), does not work very well on standard benchmarks. Presumably, the deficiency comes from the middle graph-based modules breaking the original information flow inside T5.


In order to address this problem, we present a novel architecture called \textbf{\graphix-T5} that is capable of effectively modelling relational structure information while maintaining the powerful contextual encoding capability of the pretrained T5.
\textbf{First}, we design a \graphix layer that simultaneously encodes a mixture of semantic and structural information.
Concretely, hidden states of inputs composed by questions and databases are modelled by contextualized semantic encoding, and the structural representation is injected in each transformer layer using a relational GNN block that enhances multi-hop reasoning through message passing \citep{multi-hop, gat} to capture explicit and implicit relations.
\textbf{Second}, we construct a new encoder by stacking the \graphix layers and replacing the original T5 encoder. In each \graphix layer, the parameters of the semantic block are still initialized by T5, in an attempt to maintain the contextualized encoding power of the pre-training.
In contrast to the severed GNN-T5 (Figure \ref{t5}.(c)), the \graphix-T5 (Figure \ref{t5}.(d)) will allow intensive interaction between semantic and structure from the starting layers.

We empirically show the effectiveness of \graphix-T5 on several cross-domain text-to-SQL benchmarks, \ie, \spider, \syn, \dk and \real.
On these datasets, the proposed model achieves new state-of-the-art performance, substantially outperforming all existing models by large margins. Specifically, \graphix-T5-large surprisingly beats the vanilla T5-3B.
Furthermore, we verified that \graphix-T5 can also achieve the significant improvement in the low-resource and compositional generalization obviously thanks to the introduction of structural bias.
It should be noticed that though we only focus on text-to-SQL parsing in this work, \textbf{\textit{we believe that the general methodology of \graphix-T5 can be extended to structured knowledge grounding tasks}}, \eg, TableQA \citep{wikiTQ}, Data-to-text \citep{nan-etal-2021-dart} and KBQA \citep{kgqa}.

\section{Task Formulation and Notations}
\subsection{Task Definition}
Given a natural language question $\mathcal{Q} = \left\{q_1, ..., q_{\left| \mathcal{Q} \right|}\right\}$ with its corresponding database schemas $\mathcal{D} = \left \langle \mathcal{C}, \mathcal{T} \right \rangle$, where $\mathcal{C} = \left\{c_1, ..., c_{\left| \mathcal{C} \right|}\right\}$ and $\mathcal{T} = \left\{t_1, ..., t_{\left| \mathcal{T} \right|}\right\}$ represent columns and tables, $\left| \mathcal{C} \right|$ and $\left| \mathcal{T} \right|$ refer to the number of columns and tables in each database respectively.
The goal of text-to-SQL is to generate the corresponding SQL query $y$.

\subsection{Vanilla T5 Architecture}
\subsubsection{Model Inputs}
The most canonical and effective format of inputs to T5 performing text-to-SQL task is \textit{PeteShaw} \citep{shaw-etal-2021-compositional}, which unifies natural language questions $\mathcal{Q}$ and database schema $\mathcal{D}$ as a joint sequence as shown:
\begin{equation}
\small
x = [q_{1}, ..., q_{|Q|}\,\mathrm{|} \, \mathcal{D}_{name} \, \mathrm{|} t_{1}: c_{1}^{t_{1}}, ..., c_{\left| \mathcal{C} \right|}^{t_{1}} \mathrm{|} ... \mathrm{|} t_{\left| \mathcal{T} \right|}: c_{1}^{t_{\left| \mathcal{T} \right|}}, ..., c_{\left| \mathcal{C} \right|}^{t_{\left| \mathcal{T} \right|}}| * ] ,
\label{eq1}
\end{equation}
where $q_i$ is $i^{th}$ token in the question, $t_j$ represents $j^{th}$ table in the $\mathcal{D}$, and $c_{k}^{t_{j}}$ refers to the $k^{th}$ column in the $j^{th}$ table. $*$ is the special column token in the database.
$\mathcal{D}_{name}$ is the name of each database.

\subsubsection{Encoder-Decoder Training Mechanism}
Following \citep{shaw-etal-2021-compositional}, T5 \citep{t5-jmir-2020} adopt an encoder-decoder mechanism to generate SQLs. 
First, the bi-directional encoder learns the hidden state $h$ of input $x$, then the decoder generates SQLs based on $h$ as:
\begin{equation}
\small
h =\mathbf{Enc}_{\Theta}\left(x\right); y = \mathbf{Dec}_{\Upsilon}(h) \label{eq2},
\end{equation}
where $\Theta$ and $\Upsilon$ refers to parameters of the encoder and decoder, and
$h$ connects the encoder and decoder. 
The model is initialized with pretrained T5 parameters and optimized as the following objective.
\begin{equation}
\small
\max _{\Theta,\Upsilon} \log p_{\Theta,\Upsilon}(y \mid x)=\sum_{i=1}^{|y|} \log p_{\Theta,\Upsilon}\left(y_{i} \mid y_{1: i-1}, x\right),
\label{eq4}
\end{equation}
where $x$, $y$ indicates the input and output tokens respectively and $|y|$ is the max length of generation SQL.

\section{Proposed Approach: \graphix-T5}
\subsection{Model Inputs}
\subsubsection{Contextual Encoding}
We continue to take both questions and database schemas as depicted in Eq. (\ref{eq1}) to encode the contextual information through the original T5. 

\subsubsection{Graph Construction}
The joint input questions and schemas can be displayed as a heterogeneous graph $\mathcal{G} = \left \langle \mathcal{V}, \mathcal{R} \right \rangle$ consisting of three types of nodes $\mathcal{V} = \mathcal{Q} \cup \mathcal{C} \cup \mathcal{T}$ and multiple types of relations $\mathcal{R} = \textsl{$r_1$, ..., $r_{\left| \mathcal{R} \right|}$}$, where each $r_i$ refers to a one-hop relation between nodes and a multi-hop relation $r^k$ is defined as a composition of one-hop relations: $r^k = r_{1} \circ r_{2} \cdots \circ r_{I}$ as shown in the Figure \ref{intro1}, where $I$ refers to the length of each $r^k$. 
Inspired by \citep{wang-etal-2020-rat, lgesql, sdcup,s2sql}, we enumerated a list of pre-defined relations to connect nodes. The relation sets can be divided into three main categories:
\begin{itemize}
    \item Schema relations: \textsc{Foreign-Key}, \textsc{Primary-Key}, and \textsc{Same-Table} pertain to the particular explicit schema relations that the original T5 cannot obtain from linear inputs.
    \item Schema linking relations: \textsc{Exact-Match}, \textsc{Partial-Match}, and \textsc{Value-Match} are implicit linking relations between question and schema nodes. A new type of relation \textsc{Bridge} is introduced.
    \item Question relations: \textsc{Modifier} and \textsc{Argument} are implicit dependency relations between tokens in a question. 
\end{itemize}

\subsubsection{\textsc{No-Match} Mode vs. \textsc{Bridge} Mode}
\begin{figure}
    \centering
    \includegraphics[width=0.45\textwidth]{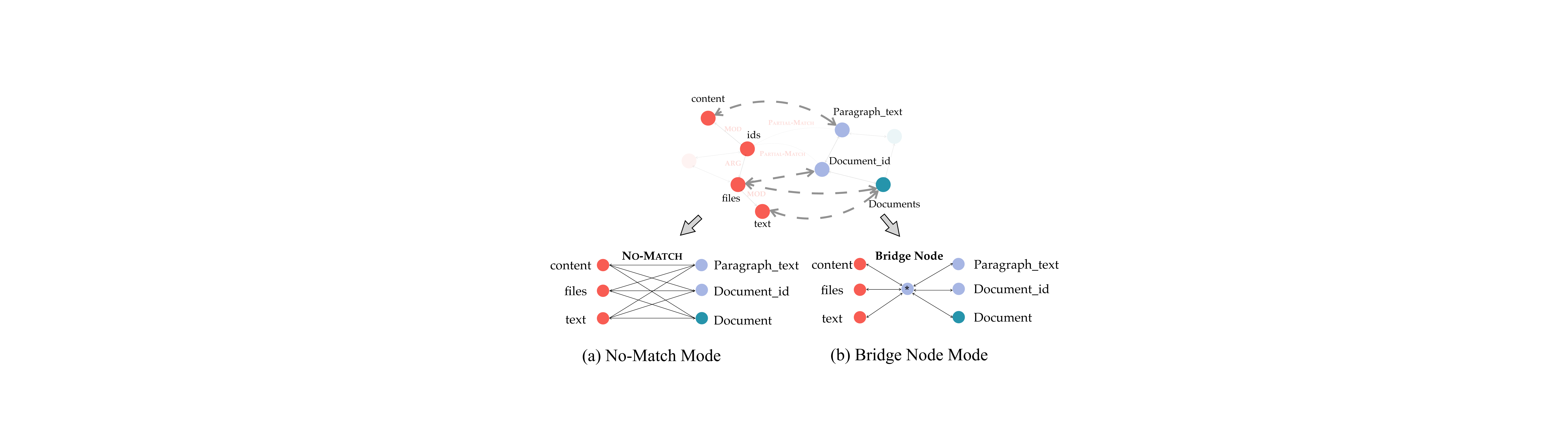}
    \caption{Figure shows the circumstances when entities in the question are hard to string-match the schema items. (a) is the strategy to solve this case by \textsc{No-Match} Mode, which fully connects schema nodes with all token nodes. (b) is our solution to add a bridge node to link the question and schema nodes via less number of edges.}
    \vspace{-0.3cm}
    \label{bridge}
\end{figure}

Previous works \citep{lgesql, s2sql} through adding the dummy edges called \textsc{No-Match} indicate that the there are question tokens and the schema tokens, which should be correlated but cannot be linked due to existing string-matched rules.
However, as shown in the Figure \ref{bridge}, \textsc{No-Match} may lead to over-smoothing problem \citep{over-smoothing} since they bring out too many noisy neighbors to compute the attention score.
Suppose there exists A tokens for the question and B schema items that are semantic relevant but not linked by the rule, the number of edges need to be linked as \textsc{No-Match} is $A \times B$.
In contrast, we leverage the special token \texttt{*} as a bridge node, allowing all schema nodes to be reached from the question token nodes by decreasing the number of edges drastically from $A \times B$ to $A + B$.

\subsection{\graphix-Layer}
The \graphix layer is designed to integrate semantic information obtained from each transformer block with structural information of a relational graph neural network (GNN) block. 
\paragraph{Semantic Representation}
The semantic representations of hidden states are firstly encoded by a Transformer \citep{transformer} block, which contains two important components, including Multi-head Self-attention Network (\textbf{MHA}) and Fully-connected Forward Network (\textbf{FFN}). In the $l^{th}$ \graphix Layer, the hidden states represent $\mathcal{H}^{l}_\mathcal{S} = \left\{h_{1}^{(l)},\ldots, h_{N}^{(l)}\right\}$, $N$ is the max length of the inputs. 
MHA at first maps query matrix $\mathbf{Q} \in \mathbb{R}^{m \times d_{k}}$, key and value matrix $\mathbf{K} \in \mathbb{R}^{n \times d_{k}}$, $\mathbf{V} \in \mathbb{R}^{n \times d_{v}}$ into an attention vector via self-attention mechanism as Eq. (\ref{eq5})
\begin{equation}
\small
\mathbf{Attn}(\mathbf{Q}, \mathbf{K}, \mathbf{V})=\mathbf{softmax}\left(\frac{\mathbf{Q} \mathbf{K}^{T}}{\sqrt{d_{k}}}\right) \mathbf{V}, \label{eq5}
\end{equation}
in which $m$ is the number of query vectors and $n$ is the number of key or value vectors. MHA executes the self-attention over $h$ heads with each head $i$ being independently parameterized by $\mathbf{W}_{i}^{Q} \in \mathbb{R}^{d_m \times d_{k}}$, $\mathbf{W}_{i}^{K} \in \mathbb{R}^{d_m \times d_{k}}$, $\mathbf{W}_{i}^{V} \in \mathbb{R}^{d_m \times d_{v}}$ and mapping inputs into queries, key-value pairs. Usually $d_{k} = d_{v} = d_{m} / h$ in the transformer blocks of T5 and $d_{m}$ denotes the dimension of T5. Then MHA calculates the attention outputs for each head and concatenate them as following:
\begin{gather}
    \small
    \mathbf{head}_{\mathrm{i}}=\mathbf{Attn}(\mathbf{Q} \mathbf{W}_{i}^{Q}, \mathbf{K} \mathbf{W}_{i}^{K}, \mathbf{V} \mathbf{W}_{i}^{V}), \\
    \mathbf{MHA}(\mathcal{H}^{(l)}_\mathcal{S})=\mathbf{Concat}\left(\mathbf{head}_{\mathrm{1}}, \cdots, \mathbf{head }_{\mathrm{h}}\right)\mathbf{W}^{O}, \\
    \widehat{\mathcal{H}}^{(l)}_\mathcal{S} = \mathbf{MHA}(\mathcal{H}^{(l)}_\mathcal{S}), \label{eq6}
\end{gather}
where $\mathbf{W}^{O} \in \mathbb{R}^{d_{m}h \times d_{m}}$ is a trainable parameter matrix.
Semantic hidden states need to be acquired through another component, \ie, FFN, which is applied as Eq. (\ref{eq8}).
\begin{equation}
\small
\mathbf{FFN}(\widehat{\mathcal{H}}^{(l)}_\mathcal{S})=\mathbf{max} \left(0,\widehat{\mathcal{H}}^{(l)}_\mathcal{S} \mathbf{W_{1}}+\mathbf{b_{1}}\right)\mathbf{W_{2}}+\mathbf{b_{2}},
\label{eq8}
\end{equation}
where linear weight matrices represent $\mathbf{W_{1}} \in \mathbb{R}^{d_{m} \times d_{ff}}$, $\mathbf{W_{2}} \in \mathbb{R}^{d_{ff} \times d_{m}}$ respectively. Experimentally, larger $d_{ff}$ is preferred, which is usually set as $d_{ff} = 4 d_{m}$. Eventually, the semantic hidden states are acquired after layer normalization and residual connection as

\begin{equation}
\small
\tilde{\mathcal{H}}^{(l)}_\mathcal{S}  = \mathbf{LayerNorm}(\widehat{\mathcal{H}}^{(l)}_\mathcal{S} + \mathbf{FFN}(\widehat{\mathcal{H}}^{(l)}_\mathcal{S})),
\label{eq12}
\end{equation}

\paragraph{Structural Representation}
In each \graphix Layer, structural representations are produced through the relational graph attention network (RGAT) \citep{rgat} over the pre-defined question-schema heterogeneous graph. Formally, given initial node embedding\footnote{\small Various initialization strategies could be implemented. In this work, we initialized the node embeddings with their semantic representations.} ${e}_{i}^{init}$  for $i^{th}$ node and its $j^{th}$ neighbor ${e}_{j}^{init}$ linked by specific types of relations, it can be computed through:

\begin{gather}
    \small
    \vec{\alpha}_{ij}=\frac{{e}_{i}^{init} \mathbf{\widetilde{W}}_{Q}\left({e}_{j}^{init} \mathbf{\widetilde{W}}_{K} + \phi \left(r_{ij}\right) \right)^{\top}}{\sqrt{d_{z}}}, \\
    \alpha_{ij}=\mathbf{softmax}_{j}\left(\vec{\alpha}_{ij} \right), \\
    \hat{e}_{i}^{init}=\sum_{j \in \mathcal{\widetilde{N}}_{i}} \alpha_{ij}\left({e}_{j}^{init} \mathbf{\widetilde{W}}_{V} + \phi(r_{ij})\right), \\
    \hat{{e}}_{i}^{(l)} = \mathbf{LayerNorm}(e_{i}^{init}+\hat{e}_{i}^{init} \mathbf{\widetilde{W}}_{O}), \\
    \tilde{{e}}_{i}^{(l)} = \mathbf{LayerNorm}(\hat{e}_{i}^{(l)}+{\mathbf{FFN}}(\hat{e}_{i}^{(l)})),
\end{gather}

Then the output node embeddings are collected as $\tilde{\mathcal{E}}^{(l)}_\mathcal{G} = \left\{\tilde{e}_{1}^{(l)}, \ldots, \tilde{e}_{N}^{(l)}\right\}$, where $\mathbf{\widetilde{W}}_{Q}$, $\mathbf{\widetilde{W}}_{K}$, $\mathbf{\widetilde{W}}_{V}$, $\mathbf{\widetilde{W}}_{O}$ $\in \mathbb{R}^{d \times d}$ are trainable parameters in the RGAT. $\phi(r_{ij})$ is a mapping function that can produce a $d$-dim embedding representing for each relation between $i^{th}$ node and $j^{th}$ node. More importantly, $\mathcal{\widetilde{N}}_{i}$ denotes the relational reception field, which is equal to the number of how many neighbors of $i^{th}$ node that RGAT will consider when updating representation of each node via message passing.


\paragraph{Jointly Representation}
After computing representations from both semantic and structural space, the $l^{th}$ \graphix Layer employs a mixture of semantic and structural information to enable information integration as following:

\begin{equation}
\small
\tilde{\mathcal{H}}^{(l)}_\mathcal{M}= \tilde{\mathcal{H}}^{(l)}_\mathcal{S} + \tilde{\mathcal{E}}^{(l)}_\mathcal{G},
\end{equation}
\subsection{\graphix-T5}
Here we present our entire \graphix-T5 model formally. The hidden states of the last layer of \graphix-encoder can be represented as:
\begin{equation}
\small
h = \mathbf{Enc}_{\Theta,\Psi}\left( x, \mathcal{G}\right), \label{eq14}
\end{equation}
where $\mathcal{G}$ is the question-schema heterogeneous graph, the $\Psi$ are the additional parameters of the RGAT, which are initialized randomly. In order to preserve the pre-trained semantic knowledge, we migrate parameters $\Theta$ from original T5 encoder as the initial parameters of semantic transformer block of the \graphix layer.


\subsection{Training}
Similar to original T5, we also follow a fine-tuning strategy. The whole training framework is to optimize the following log-likelihood.
\begin{equation}
\small
    \max _{\Theta,\Upsilon,\Psi} \log p_{\Theta,\Upsilon,\Psi}(y \mid x) =\sum_{i=1}^{|y|} \log p_{\Theta,\Upsilon,\Psi}\left(y_{i} \mid y_{1: i-1}, x, \mathcal{G}\right).
\end{equation}

\section{Experiment}

\subsection{Set up}

\paragraph{Datasets and Settings}
We conduct extensive experiments on four challenging benchmarks for cross-domain text-to-SQLs and two different training settings. 
\textbf{(1) \textsc{Spider}} \citep{yu-etal-2018-spider} is a large-scale cross-domain text-to-SQL benchmark, also including 9 previous classic datasets, \eg, Scholar \citep{scholar}, WikiSQL \citep{wikiSQL}, GeoQuery \citep{geoquery}, etc. 
It contains 8659 training examples and 1034 development examples, which covers 200 complex databases across 138 domains. The testing set is not available for individual review.
\textbf{(2) \textsc{Syn}} \citep{Syn} replaces the simple string-matched question tokens or schema names with their synonyms.
\textbf{(3) \textsc{Dk}} \citep{DK} requires the text-to-SQL parsers to equip with the capability of domain knowledge reasoning.
\textbf{(4) \textsc{Realistic}} removes and switches the obvious mentions of schema items in questions, making it closer to the real scenarios. 
Furthermore, we also test the compositional generalization ability of our model on the
\textbf{\textsc{Spider-SSP}} \citep{shaw-etal-2021-compositional} with three splits from \textsc{Spider}: Spider-Length (split dataset based on variant lengths); Spider-TMCD (Target Maximum Compound Divergence) and Spider-Template (split based on different parsing templates). Finally, the performances of \graphix-T5 on \textbf{\textsc{Low-Resource}} setting are evaluated on usage of 10\%, 20\%, 50\% data separately.

\begin{table}[t]  
    \centering
    \resizebox{0.8\hsize}{!}{
    \begin{tabular}{lcc}  
    \toprule
    \textbf{\textsc{Model}}& \textbf{\textsc{Em}} & \textbf{\textsc{Ex}} \\ 
    \midrule
    RAT-SQL + BERT  $^\heartsuit$       & 69.7 & - \\
    RAT-SQL + Grappa $^\heartsuit$      & 73.9 & - \\
    GAZP + BERT                         & 59.1 &59.2\\ 
    BRIDGE v2 + BERT                    & 70.0 &68.3\\
    NatSQL + GAP                          & 73.7 &75.0 \\
    SMBOP + GRAPPA                      & 74.7 &75.0\\ 
    LGESQL + ELECTRA $^\heartsuit$      & 75.1 & - \\
    S$^2$SQL + ELECTRA $^\heartsuit$    & 76.4 & - \\
    \midrule
    \midrule
    T5-large                                                                & 67.0                              & 69.3 \\
    \rowcolor[RGB]{237,237,237} \graphix-T5-large                            & 72.7$_\textbf{($\uparrow$ 5.7)}$    & 75.9$_\textbf{($\uparrow$ 6.6)}$ \\
    T5-large + \picard $^\clubsuit$                                         & 69.1                              & 72.9 \\
    \rowcolor[RGB]{237,237,237} \graphix-T5-large + \picard $^\clubsuit$     & 76.6$_\textbf{($\uparrow$ 7.5)}$    & 80.5$_\textbf{($\uparrow$ 7.6)}$\\
    \midrule
    T5-3B                                                                   & 71.5                              & 74.4 \\
    \rowcolor[RGB]{237,237,237} \graphix-T5-3B                               & {75.6}$_\textbf{ ($\uparrow$ 4.1)}$  & {78.2}$_\textbf{ ($\uparrow$ 3.8)}$ \\
    T5-3B + \picard $^\clubsuit$                                            & 75.5                              & 79.3\\
    \rowcolor[RGB]{237,237,237} \graphix-T5-3B + \picard $^\clubsuit$        & \textbf{77.1}$_\textbf{($\uparrow$ 1.6)}$       & \textbf{81.0}$_\textbf{($\uparrow$ 1.7)}$ \\
    \bottomrule
    \end{tabular}}
    \caption{Exact match ({EM}) and execution  ({EX}) accuracy (\%) on \spider development set. $^\heartsuit$ means the model does not predict SQL values. $^\clubsuit$ means the model uses the constrained decoding \picard. $\uparrow$ is an absolute improvement.}
    \label{tab:spider}
    \vspace{-0.4cm}
\end{table}

\paragraph{Evaluation Metrics}
Following \citep{yu-etal-2018-spider}, Exact Match (EM) and Execution Accuracy (EX) are the two standard metrics we use to measure performance of our model. EM can evaluate how much a generated SQL is comparable to the gold SQL. EX can reflect whether a predicted SQL is valid and returns the exact result as desired by users. 

\paragraph{Implementation Details}
We implement our codes \footnote{\url{https://github.com/AlibabaResearch/DAMO-ConvAI/tree/main/graphix}} mainly based on  hugging-face transformers library \citep{hf-transformers} \footnote{https://huggingface.co/}. We set the max input length as 1024, generation max length as 128, and batch size as 32. We also adopt Adafactor \citep{adafactor} as our primary optimizer with a linear decayed learning rate of 5e-5.  During the experiment, \graphix layers are mainly injected into the encoder to learn better representations for structural generalization. We evaluate our effectiveness of \graphix-T5 across two main versions: T5-Large with approximately 800M parameters and T5-3B, with more than 3 Billion parameters literally. All experiments are conducted on one NVIDIA Tesla A100, which is available for the most research centers.

\paragraph{Compared Methods}
Our model are compared mainly to mainstream strong baseline models such as GNNSQL \citep{GNNSQL}, RATSQL \citep{wang-etal-2020-rat}, GAZP \citep{GAZP}, BRIDEGE \citep{Bridging}, SMBOP \citep{smbop}, NatSQL \citep{natsql}, LGESQL \citep{lgesql}, S$^2$SQL \citep{s2sql} and T5+PICARD \citep{PICARD} across the disparate datasets and settings.

\begin{table}[t]  
    \centering
    \resizebox{0.8\hsize}{!}{
    \begin{tabular}{lccc}  
    \toprule
    \textbf{\textsc{Model}}& \textbf{\syn} &  \textbf{\dk}& \textbf{\real} \\ 
    \midrule
    GNN & 23.6 & 26.0 & - \\
    IRNet &28.4 &33.1 & - \\
    RAT-SQL & 33.6 &35.8 & - \\
    RAT-SQL + BERT & 48.2 &40.9 & 58.1 \\
    RAT-SQL + Grappa & 49.1 &38.5 & 59.3 \\
    LGESQL + ELECTRA & 64.6 &48.4 & 69.2 \\
    \midrule
    \midrule
    T5-large & 53.6  &40.0 & 58.5 \\
    \rowcolor[RGB]{237,237,237} \graphix-T5-large & 61.1$_\textbf{ ($\uparrow$ 7.5)}$ & 48.6$_\textbf{ ($\uparrow$ 8.6)}$ & 67.3$_\textbf{ ($\uparrow$ 8.8)}$ \\
    
    \midrule
    T5-3B & 58.0  &46.9& 62.0 \\
    \rowcolor[RGB]{237,237,237} \graphix-T5-3B & $\textbf{66.9}_\textbf{ ($\uparrow$ 8.9)}$ & $\textbf{51.2}_\textbf{ ($\uparrow$ 4.3)}$ & $\textbf{72.4}_\textbf{ ($\uparrow$ 10.4)}$ \\
    \bottomrule
    \end{tabular}}
    \caption{Exact match ({EM}) accuracy (\%) on \syn, \dk and \real benchmark.}
    \label{tab:robust}
\end{table}

\begin{table}[t]  
    \centering
    \resizebox{0.8\hsize}{!}{
    \begin{tabular}{lccc}  
    \toprule
    \textbf{\textsc{Model}}& \textbf{\textsc{Template}} & \textbf{\textsc{Length}} & \textbf{\textsc{Tmcd}} \\ 
    \midrule
    T5-base     & 59.3 & 49.0 & 60.9 \\
    T5-3B       & 64.8 & 56.7 & 69.6 \\
    NQG-T5-3B   & 64.7 & 56.7 & 69.5 \\
    \midrule
    \rowcolor[RGB]{237,237,237} \graphix-T5-3B & $\textbf{70.1}_\textbf{ ($\uparrow$ 5.4)}$  & $\textbf{60.6}_\textbf{ ($\uparrow$ 3.9)}$ & $\textbf{73.8}_\textbf{ ($\uparrow$ 4.3)}$\\
    \bottomrule
    \end{tabular}}
    \caption{Exact match ({EM}) accuracy (\%) on compositional dataset \textsc{Spider-Ssp}.}
    \label{tab:ssp}
    \vspace{-0.4cm}
\end{table}

\begin{table*}[t]
\centering
\resizebox{1\hsize}{!}{
\begin{tabular}{l|ccccc|ccccc|ccccc|ccccc}
\toprule
{\multirow{2}*{\textbf{\textsc{Model}}}} & \multicolumn{5}{c|}{\textbf{\spider}}      & \multicolumn{5}{c|}{\textbf{\syn}}      & \multicolumn{5}{c}{\textbf{\dk}}  & \multicolumn{5}{c}{\textbf{\real}}   \\
        & easy & medium & hard & extra & all & easy & medium & hard & extra & all & easy & medium & hard & extra & all & easy & medium & hard & extra & all \\
\midrule
T5-large & 85.5& 70.9& 55.2&41.6&67.0       &69.0&56.8&46.3&30.2&53.6     &\textbf{64.1}&44.3&22.9&18.1&40.0    &79.8&68.0&44.4&28.9&58.5  \\
\rowcolor[RGB]{237,237,237} \graphix-T5-large &\textbf{89.9}&\textbf{78.7}&\textbf{59.8}&\textbf{44.0}&\textbf{72.6}      &\textbf{75.8}&\textbf{67.5}&\textbf{50.6} &\textbf{33.1}&\textbf{61.1}  &63.6&\textbf{54.5}&\textbf{33.8}&\textbf{29.5}&\textbf{48.6}   &\textbf{88.1}&\textbf{77.3} &\textbf{50.5}&\textbf{40.2}&\textbf{67.3}  \\
\midrule
T5-3B    &89.5&78.3 &58.6&40.4&71.6     &74.2 &64.5&48.0&27.8&58.0   &\textbf{69.9}&53.5&24.3&24.8&46.9    &85.3&73.4&46.5&27.8&62.0 \\
\rowcolor[RGB]{237,237,237} \graphix-T5-3B & \textbf{91.9}& \textbf{81.6}& \textbf{61.5}&\textbf{50.0} &\textbf{75.6} &  \textbf{80.6}& \textbf{73.1}&\textbf{52.9}& \textbf{44.6} & \textbf{66.9}   &69.1&\textbf{55.3}&\textbf{39.2}&\textbf{31.4}&\textbf{51.2}    &\textbf{93.6} &\textbf{85.7} &\textbf{52.5} & \textbf{41.2} & \textbf{72.4} \\
\bottomrule
\end{tabular}}
\caption{Exact matching (EM) accuracy by varying the levels of difficulty of the inference data on four benchmarks.}
\label{tab:diff_result}
\vspace{-0.3cm}
\end{table*}

\subsection{Overall Performance}

\paragraph{Results on \spider}
Table \ref{tab:spider} displays the performance of \graphix-T5 and other competitive baseline models on official \textsc{Spider} benchmark. First, we demonstrate that \graphix-T5-3B with a constrained decoding module PICARD \citep{PICARD} achieves the state-of-the-art on this challenging cross-domain text-to-SQL benchmark. Also, it is evident that \graphix-T5 is vastly superior to the vanilla T5 on large and 3B scale with a significant margin. This indicates that the structural generalization capability of the \graphix layer is crucial for T5 such a text-to-text PLM to perform the text-to-SQL task. 

\paragraph{Zero-shot Results on More Challenging Settings}
As shown in the Table \ref{tab:robust}, we further demonstrate the robustness of \graphix-T5 when it confronts with more challenging and closer to realistic evaluations in \textsc{Syn}, \textsc{Dk}, \textsc{Realistic} without any additional training. First of all, the results show that \graphix-T5-3B outperforms other baseline models across all three datasets. Furthermore, we observe that \graphix-T5-large and \graphix-T5-3B surpass the performance of vanilla T5-large and T5-3B with a clear margin, respectively. This demonstrates that vanilla T5 is hungry for structural reasoning when dealing with more flexible and complicated questions for text-to-SQLs from real-world scenarios. And \graphix can mitigate this problem.
\begin{figure}
    \centering
    \includegraphics[width=0.45\textwidth]{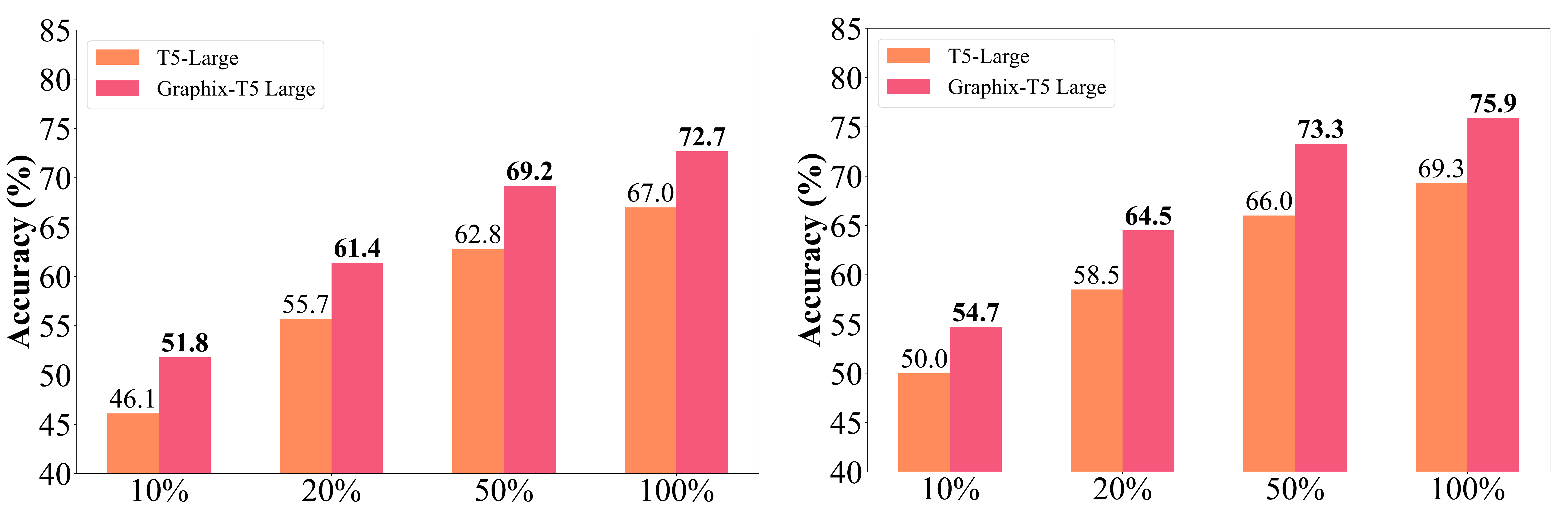}
    \caption{Exact match ({EM}) (left) and execution ({EX}) (right) accuracy (\%) on \spider low-resource setting.}
    \label{few}
    \vspace{-0.4cm}
\end{figure}

\paragraph{Results on Compositional Generalization}
As shown in Table \ref{tab:ssp}, on \textsc{Spider-Ssp}, the grammar-based inductive T5 model provided by \citep{shaw-etal-2021-compositional}, named NQG-T5, has no obvious advantages over vanilla T5, which indicates that the grammar of natural language is not helpful to enhance T5 for compositional generation. However, \graphix-T5 helps the T5 gain the SQL knowledge and makes it less vulnerable to these modifications through the effective fusion of structural information.

\paragraph{Results on Low-resource Settings}
Figure \ref{few} records the performance of \graphix-T5-large and T5-large on different low-resource settings.
It displays 1) in each low-resource setting, \graphix-T5-large performs considerably better than vanilla T5-large. It demonstrates that the structural knowledge created by humans can compensate for the inadequate learning due to low-resource data \citep{ontology}; 2) notably, \graphix-T5-large can perform obviously better than the vanilla T5-large trained on \textbf{100\%} data even within just usage of \textbf{50\%} data. This further verifies the strengths of \graphix-T5 for training in the low-data resources.
\paragraph{Results on Complex Queries}
As presented in Table \ref{tab:diff_result}, we also compare the more precise performance results of \graphix-T5 to the vanilla T5 in four separate SQL difficulty levels splitted by \textsc{Spider} officially, in order to better comprehend the performance improvements. We observe that \graphix-T5 is more capable of handling harder text-to-SQL cases, as illustrated in the \textbf{Hard} and \textbf{Extra-hard} examples, indicating that structural bias training is beneficial 
to the text-to-text PLMs to reason over complex scenarios.

\begin{table}[t]  
    \centering
    \setlength\tabcolsep{12pt}{
    \resizebox{0.65\hsize}{!}{
    \begin{tabular}{lcc}  
    \toprule
    \textbf{\textsc{Model}}& \textbf{\textsc{Em}} & \textbf{\textsc{Ex}} \\ 
    \midrule
    (a) RAT-SQL + BERT & 69.7 & - \\
    \midrule
    (b) T5-large & 67.0 & 69.3 \\
    \midrule 
    (c) GNN-T5-large     & 51.6 & 54.5 \\
    \midrule
    \midrule
    (d) \graphix-T5-large \\
    \quad w/ \textsc{Bridge} Mode    & \textbf{72.7} & \textbf{75.9} \\
    \quad w/ \textsc{No-Match} Mode    & 71.1 & 74.2 \\
    \quad w/ \textsc{Double-Graph} & 72.0 & 74.7 \\
    \bottomrule
    \end{tabular}
    }
    }
    \caption{Ablation study for the variant GNN + PLM tactics on cross-domain text-to-SQLs, echoing Figure \ref{t5}, (a) is RATSQL, (b) is vanilla T5, (c) is GNN-T5 and (d) is \graphix.}
    \label{tab:ablation}
    \vspace{-0.4cm}
\end{table}

\begin{figure}
    \centering
    \includegraphics[width=0.3\textwidth]{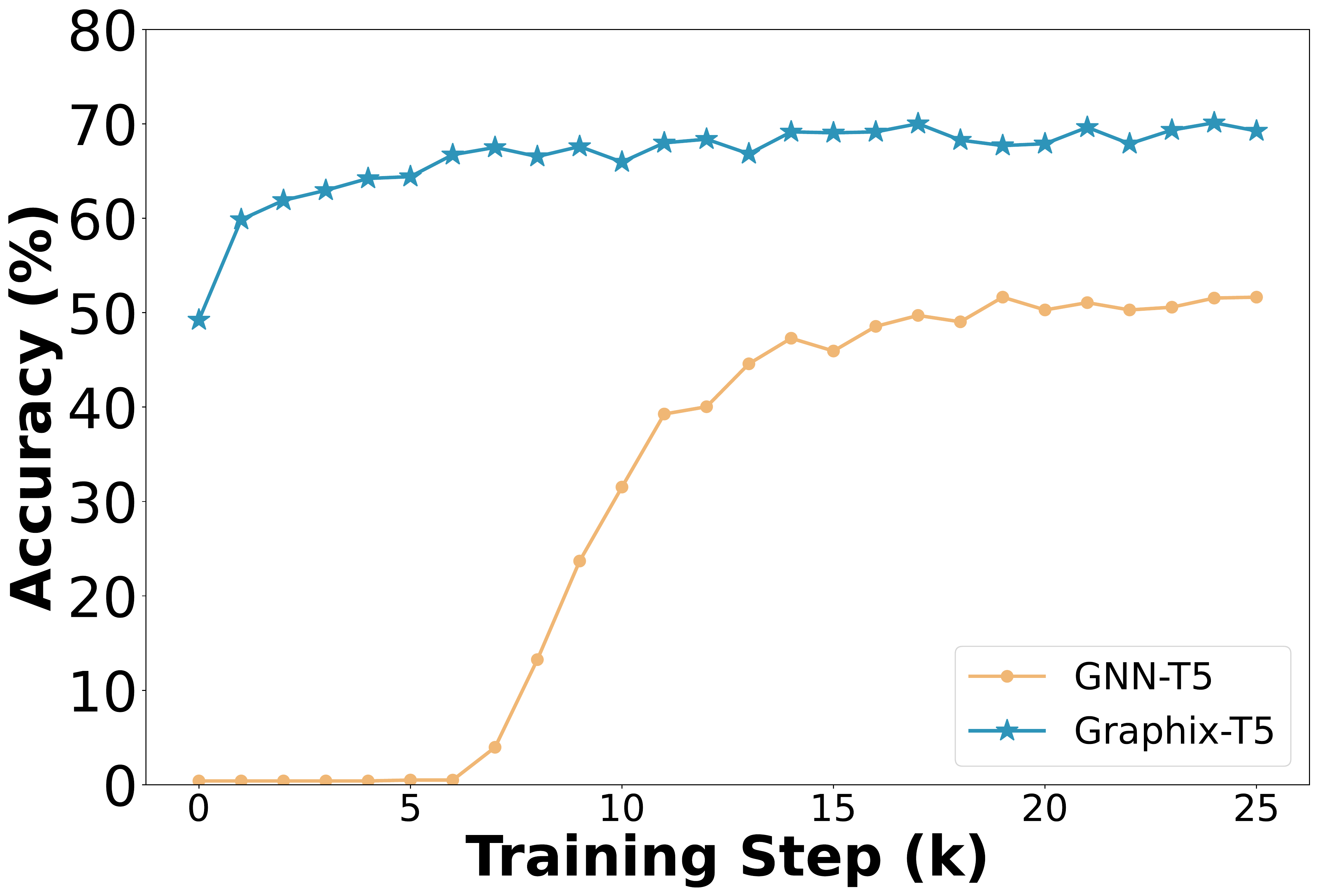}
    \caption{The performance of the validation sets during the convergence of \graphix-T5 and GNN-T5 on \spider. It can be clearly demonstrated that GNN-T5 has extremely unsatisfactory performance, due to catastrophic forgetting.}
    \label{training}
    \vspace{-0.4cm}
\end{figure}

\begin{figure*}
    \centering
    \includegraphics[width=0.7\textwidth]{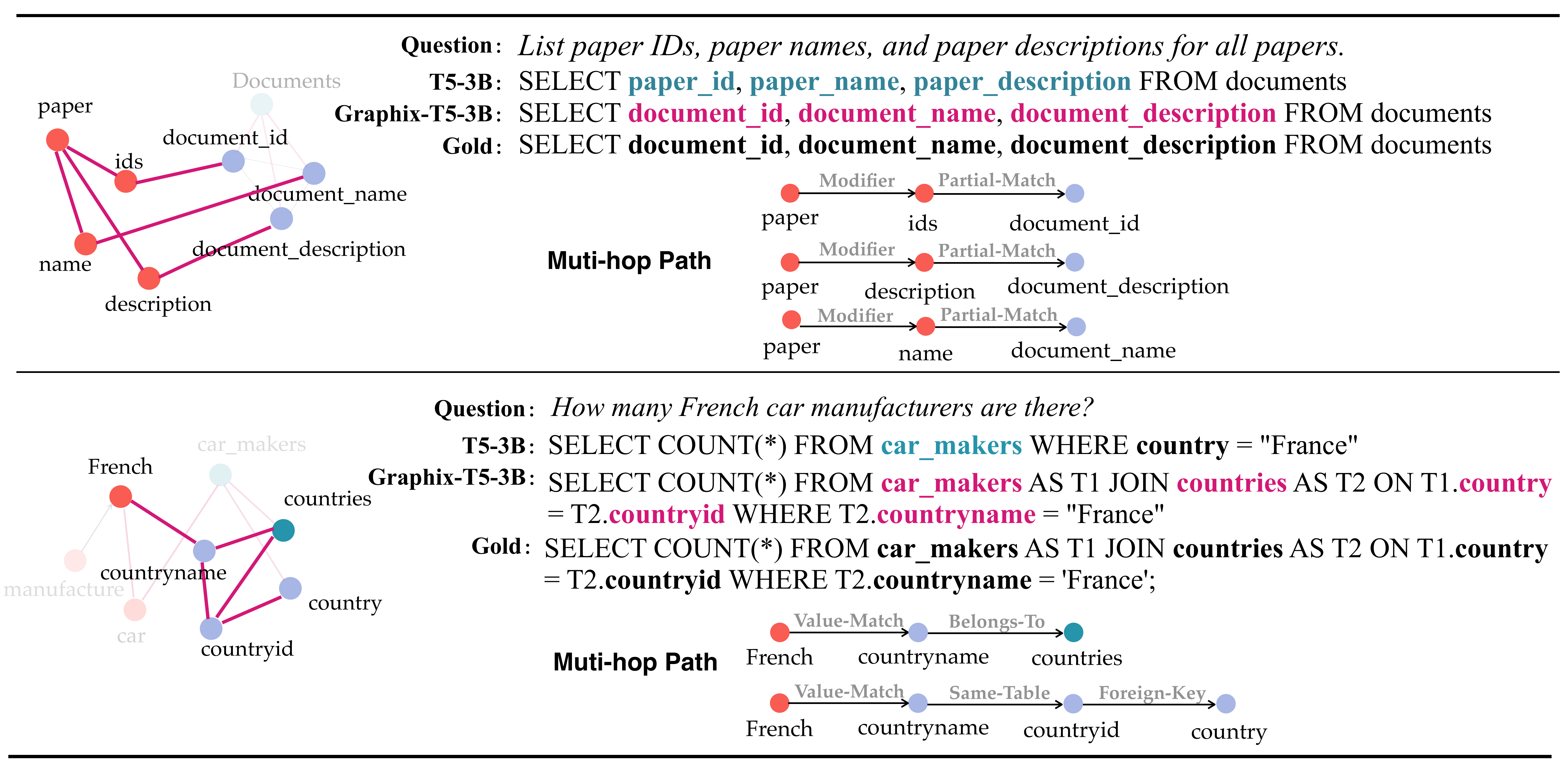}
    \caption{Case study: two illustrative cases sampled randomly from \textsc{Syn}. It shows that multi-hop reasoning can help \graphix-T5 generate more correct SQLs in terms of semantic meanings and database schema structures.}
    \label{case}
    \vspace{-0.3cm}
\end{figure*}

\subsection{Ablation Study}
As shown in Table \ref{tab:ablation}, to better validate the function of each component of \graphix-T5, ablation studies are performed in large version and expected to answer the following questions.
\paragraph{\texttt{[1]} How effective is \textsc{Bridge Mode} ?} \graphix-T5-large with \textsc{Bridge Mode} can achieve the better performance than with \textsc{No-Match} Mode. It indicates that \textsc{No-match} mode will greatly increase the number of noisy neighbors, resulting in higher risk of over-smoothing issues \citep{over-smoothing}.

\paragraph{\texttt{[2]} Could \graphix be incorporated into decoder ?} With \textsc{Double-Graph} means that \graphix-T5 incorporate \graphix layer into the both encoder and decoder. The result reveals that adding \graphix layers to the decoder does not lead to any improvements.  Since decoder is an auto-regressive model, which only considers the history tokens when generating the current token. However, \graphix-T5, which can forecast the information of future tokens by global linking, may disrupt this characteristic leading to the negative impact on the decoder. Therefore, we propose that the best tactic is to only incorporate \graphix layers into the encoder.

\paragraph{\texttt{[3]} Is \graphix superior than other architecture variants ?} 
Echoing Figure \ref{t5}, we access the performance of 4 categories of models using PLMs on \spider. 
According to Table \ref{tab:ablation} (c), the performance of GNN-T5 has decreased by roughly 20\% when compared to \graphix-T5, proving GNN-T5 training strategy to be ineffective. 
Moreover, we notice that such severed GNN-T5 encounters a catastrophic forgetting problem \citep{catastrophic} during training. 
Since the accuracy of the GNN-T5 continues to be 0 in the first thousands of steps, as shown in Figure \ref{training}, it is evident that all pretrained knowledge from T5 would be forgotten. 
After convergence, the GNN-T5 performance decreases significantly from the \graphix-T5, indicating that only a small portion of the semantic information from T5 has been utilized. 
In contrast, \graphix-T5 can achieve almost 50\% accuracy inside the first 1000 training steps and more than 20\% improvement than GNN-T5 after convergence, which verifies the advantages of \graphix-T5 that can avoid catastrophic forgetting and augment generalization capability.

\subsection{Case Study}
To illustrate the effectiveness of \graphix qualitatively, two examples are displayed in Figure \ref{case}, which are sampled randomly from \textsc{Syn}. Figure \ref{case} shows the comparison of predicted SQLs by vanilla T5-3B and \graphix-T5-3B. We can observe that \graphix can generate correct SQLs even in the hard scenarios. That is because that, even with a small number of keywords overlapped, \graphix-T5 can accurately identify counterpart column or table objects and generate a high-quality SQL through multi-hop reasoning and structural grounding. For example, in the first case, vanilla T5-3B picks the incorrect columns \texttt{paper\_id}, \texttt{paper\_name}, and \texttt{paper\_description}, which even don't appear in the table documents. This implies that vanilla T5-3B is unable to reach the target schema elements without the capability of structural grounding when confronting challenging text-to-SQLs. Instead, \graphix-T5-3B can correspond the question entities to the correct column names through multi-hop paths presented in the Figure \ref{case}. In the second case, vanilla T5-3B misidentifies the \texttt{country} as their target column, however, \texttt{"France"} only appears in the column \texttt{countryname} of the table \texttt{countries}. This suggests T5-3B is only able to generate semantically valid SQLs, which fails to take into account the real database structure. On contrary, \graphix-T5 can produce truly valid SQLs in terms of both questions and databases via a successful mixture of semantic and structural information during training.

\section{Related Works}
The basic principle of a cross-domain text-to-SQL parser is to build an encoder to learn the representations of the questions and schemas, while employing a decoder to generate SQLs with the information learnt in the encoder \citep{t2s_survey}. In particular, IRNET \citep{irnet} proposes to design an encoder to learn the representations of questions and schemas respectively via an attention-based Bi-LSTM and a decoder to predict SQLs via encoded intermediate representations. 
Later, the graph-based encoders have been successfully proved its effectiveness in text-to-SQL tasks, for example, some works \citep{GNNSQL, chen-etal-2021-shadowgnn} construct the schema graph and enhance the representations of inputs. RATSQL \citep{wang-etal-2020-rat}, SDSQL \citep{sdsql}, LGESQL \citep{lgesql}, S$^2$SQL \citep{s2sql} further improve structural reasoning through modelling relations between schema and questions. R$^2$SQL \citep{r2sql}, \textsc{Score} \citep{score} and \textsc{Star} \citep{star} enhance structural reasoning for context-dependent text-to-SQL parsing.
These works are performed by the PLM independently building the semantic features, followed by the graph-based module injecting the structural information.
However, such training strategy is just effective to encoder-based PLMs (\ie, BERT \citep{devlin-etal-2019-bert}, ELECTRA \citep{electra-iclr-2020}, \etal).

Recently, the text-to-text PLM T5 has been proven effectiveness in text-to-SQL \citep{shaw-etal-2021-compositional, sun}. Besides, \citep{PICARD} designs a constrained decoding process, namely PICARD, to detect and refuse erroneous tokens during the beam-search phase.
\citet{UnifiedSKG} further injects the knowledge from other structural knowledge grounding tasks into T5 with multi-task to boost performance on text-to-SQL.  
Despite effectiveness, these methods still struggle to generate SQLs in the more challenging and complex scenarios without explicit and implicit structural information.
However, \graphix-T5 can overcome this issue by an argument of graph representation learning in the encoder. Concurrently, RASAT \citep{rasat} also attempts to provide T5 with the structural information by adding edge embedding into the multi-head self-attention, while we keep the pre-trained transformers complete in order to benefit the most from prior semantic knowledge, which leads to better performance.

\section{Conclusion}
In this paper, we proposed an effective architecture to boost the capability of structural encoding of T5 cohesively while keeping the pretrained T5's potent contextual encoding ability. In order to achieve this goal, we designed a Graph-Aware semi-pretrained text-to-text PLM, namely \graphix-T5, to augment the multi-hop reasoning for the challenging text-to-SQL task. The results under the extensive experiments demonstrate the effectiveness of \graphix-T5, proving that structural information is crucial for the current text-to-text PLMs for complicated text-to-SQL cases.

\section*{Acknowledgement}
We thank Dr. Tao Yu and Tianbao Xie for evaluation of our work on SPIDER leaderboard.
We thank Dr. Bailin Wang and Dr. Bowen Li for constructive suggestions.
Reynold Cheng, Jinyang Li, Nan Huo, and Wenyu Du were supported by the University of Hong Kong (Project 104006830), the Guangdong–Hong Kong-Macau Joint Laboratory Program 2020 (Project No: 2020B1212030009), and the Innovation Wing Two Research fund.
Jinyang Li was also supported by HKU Presidential PhD Scholar Programme and Alibaba Group through Alibaba Research Intern Program. Chenhao Ma was supported in part by Shenzhen Science and Technology Program under grant No.ZDSYS20211021111415025.
\bibliography{aaai23.bib}

\begin{thebibliography}{48}
\providecommand{\natexlab}[1]{#1}

\bibitem[{Bogin, Berant, and Gardner(2019)}]{GNNSQL}
Bogin, B.; Berant, J.; and Gardner, M. 2019.
\newblock Representing Schema Structure with Graph Neural Networks for
  Text-to-{SQL} Parsing.
\newblock In \emph{Proc. of ACL}.

\bibitem[{Cai et~al.(2018)Cai, Xu, Zhang, Yang, Li, and Liang}]{ijcai2018-553}
Cai, R.; Xu, B.; Zhang, Z.; Yang, X.; Li, Z.; and Liang, Z. 2018.
\newblock An Encoder-Decoder Framework Translating Natural Language to Database
  Queries.
\newblock In \emph{Proc. of IJCAI}.

\bibitem[{Cai et~al.(2021)Cai, Yuan, Xu, and Hao}]{sadga}
Cai, R.; Yuan, J.; Xu, B.; and Hao, Z. 2021.
\newblock {SADGA:} Structure-Aware Dual Graph Aggregation Network for
  Text-to-SQL.
\newblock In \emph{Proc. of NeurIPS}.

\bibitem[{Cai et~al.(2022)Cai, Li, Hui, Yang, Li, Li, Cao, Li, Huang, Si, and
  Li}]{star}
Cai, Z.; Li, X.; Hui, B.; Yang, M.; Li, B.; Li, B.; Cao, Z.; Li, W.; Huang, F.;
  Si, L.; and Li, Y. 2022.
\newblock STAR: SQL Guided Pre-Training for Context-dependent Text-to-SQL
  Parsing.
\newblock In \emph{Proc. of EMNLP Findings}.

\bibitem[{Cao et~al.(2021)Cao, Chen, Chen, Zhao, Zhu, and Yu}]{lgesql}
Cao, R.; Chen, L.; Chen, Z.; Zhao, Y.; Zhu, S.; and Yu, K. 2021.
\newblock {LGESQL}: Line Graph Enhanced Text-to-{SQL} Model with Mixed Local
  and Non-Local Relations.
\newblock In \emph{Proc. of ACL}.

\bibitem[{Chen et~al.(2020{\natexlab{a}})Chen, Lin, Li, Li, Zhou, and
  Sun}]{over-smoothing}
Chen, D.; Lin, Y.; Li, W.; Li, P.; Zhou, J.; and Sun, X. 2020{\natexlab{a}}.
\newblock Measuring and Relieving the Over-Smoothing Problem for Graph Neural
  Networks from the Topological View.
\newblock In \emph{Proc. of AAAI}.

\bibitem[{Chen et~al.(2020{\natexlab{b}})Chen, Meng, Li, Chen, Xu, Xu, and
  Zhou}]{Bridging}
Chen, X.; Meng, F.; Li, P.; Chen, F.; Xu, S.; Xu, B.; and Zhou, J.
  2020{\natexlab{b}}.
\newblock Bridging the Gap between Prior and Posterior Knowledge Selection for
  Knowledge-Grounded Dialogue Generation.
\newblock In \emph{Proc. of EMNLP}.

\bibitem[{Chen et~al.(2021)Chen, Chen, Zhao, Cao, Xu, Zhu, and
  Yu}]{chen-etal-2021-shadowgnn}
Chen, Z.; Chen, L.; Zhao, Y.; Cao, R.; Xu, Z.; Zhu, S.; and Yu, K. 2021.
\newblock {S}hadow{GNN}: Graph Projection Neural Network for Text-to-{SQL}
  Parser.
\newblock In \emph{Proc. of NAACL}.

\bibitem[{Clark et~al.(2020)Clark, Luong, Le, and Manning}]{electra-iclr-2020}
Clark, K.; Luong, M.; Le, Q.~V.; and Manning, C.~D. 2020.
\newblock {ELECTRA:} Pre-training Text Encoders as Discriminators Rather Than
  Generators.
\newblock In \emph{Proc. of ICLR}.

\bibitem[{Devlin et~al.(2019)Devlin, Chang, Lee, and
  Toutanova}]{devlin-etal-2019-bert}
Devlin, J.; Chang, M.-W.; Lee, K.; and Toutanova, K. 2019.
\newblock {BERT}: Pre-training of Deep Bidirectional Transformers for Language
  Understanding.
\newblock In \emph{Proc. of NAACL}.

\bibitem[{Fang et~al.(2020)Fang, Sun, Gan, Pillai, Wang, and Liu}]{multi-hop}
Fang, Y.; Sun, S.; Gan, Z.; Pillai, R.; Wang, S.; and Liu, J. 2020.
\newblock Hierarchical Graph Network for Multi-hop Question Answering.
\newblock In \emph{Proc. of EMNLP}.

\bibitem[{French(1999)}]{catastrophic}
French, R.~M. 1999.
\newblock Catastrophic forgetting in connectionist networks.
\newblock \emph{Trends in cognitive sciences}.

\bibitem[{Gan et~al.(2021{\natexlab{a}})Gan, Chen, Huang, Purver, Woodward,
  Xie, and Huang}]{Syn}
Gan, Y.; Chen, X.; Huang, Q.; Purver, M.; Woodward, J.~R.; Xie, J.; and Huang,
  P. 2021{\natexlab{a}}.
\newblock Towards Robustness of Text-to-{SQL} Models against Synonym
  Substitution.
\newblock In \emph{Proc. of ACL}.

\bibitem[{Gan, Chen, and Purver(2021)}]{DK}
Gan, Y.; Chen, X.; and Purver, M. 2021.
\newblock Exploring Underexplored Limitations of Cross-Domain Text-to-{SQL}
  Generalization.
\newblock In \emph{Proc. of EMNLP}.

\bibitem[{Gan et~al.(2021{\natexlab{b}})Gan, Chen, Xie, Purver, Woodward,
  Drake, and Zhang}]{natsql}
Gan, Y.; Chen, X.; Xie, J.; Purver, M.; Woodward, J.~R.; Drake, J.; and Zhang,
  Q. 2021{\natexlab{b}}.
\newblock Natural {SQL}: Making {SQL} Easier to Infer from Natural Language
  Specifications.
\newblock In \emph{Proc. of EMNLP Findings}.

\bibitem[{Guo et~al.(2019)Guo, Zhan, Gao, Xiao, Lou, Liu, and Zhang}]{irnet}
Guo, J.; Zhan, Z.; Gao, Y.; Xiao, Y.; Lou, J.-G.; Liu, T.; and Zhang, D. 2019.
\newblock Towards Complex Text-to-{SQL} in Cross-Domain Database with
  Intermediate Representation.
\newblock In \emph{Proc. of ACL}.

\bibitem[{Hui et~al.(2021{\natexlab{a}})Hui, Geng, Ren, Li, Li, Sun, Huang, Si,
  Zhu, and Zhu}]{r2sql}
Hui, B.; Geng, R.; Ren, Q.; Li, B.; Li, Y.; Sun, J.; Huang, F.; Si, L.; Zhu,
  P.; and Zhu, X. 2021{\natexlab{a}}.
\newblock Dynamic Hybrid Relation Network for Cross-Domain Context-Dependent
  Semantic Parsing.
\newblock In \emph{Proc. of AAAI}.

\bibitem[{Hui et~al.(2022)Hui, Geng, Wang, Qin, Li, Li, Sun, and Li}]{s2sql}
Hui, B.; Geng, R.; Wang, L.; Qin, B.; Li, Y.; Li, B.; Sun, J.; and Li, Y. 2022.
\newblock S{\({^2}\)}SQL: Injecting Syntax to Question-Schema Interaction Graph
  Encoder for Text-to-SQL Parsers.
\newblock In \emph{Proc. of ACL Findings}.

\bibitem[{Hui et~al.(2021{\natexlab{b}})Hui, Shi, Geng, Li, Li, Sun, and
  Zhu}]{sdsql}
Hui, B.; Shi, X.; Geng, R.; Li, B.; Li, Y.; Sun, J.; and Zhu, X.
  2021{\natexlab{b}}.
\newblock Improving Text-to-SQL with Schema Dependency Learning.
\newblock In \emph{arXiv:2103.04399}.

\bibitem[{Iyer et~al.(2017)Iyer, Konstas, Cheung, Krishnamurthy, and
  Zettlemoyer}]{scholar}
Iyer, S.; Konstas, I.; Cheung, A.; Krishnamurthy, J.; and Zettlemoyer, L. 2017.
\newblock Learning a Neural Semantic Parser from User Feedback.
\newblock In \emph{Proc. of ACL}.

\bibitem[{Nan et~al.(2021)Nan, Radev, Zhang, Rau, Sivaprasad, Hsieh, Tang,
  Vyas, Verma, Krishna, Liu, Irwanto, Pan, Rahman, Zaidi, Mutuma, Tarabar,
  Gupta, Yu, Tan, Lin, Xiong, Socher, and Rajani}]{nan-etal-2021-dart}
Nan, L.; Radev, D.; Zhang, R.; Rau, A.; Sivaprasad, A.; Hsieh, C.; Tang, X.;
  Vyas, A.; Verma, N.; Krishna, P.; Liu, Y.; Irwanto, N.; Pan, J.; Rahman, F.;
  Zaidi, A.; Mutuma, M.; Tarabar, Y.; Gupta, A.; Yu, T.; Tan, Y.~C.; Lin,
  X.~V.; Xiong, C.; Socher, R.; and Rajani, N.~F. 2021.
\newblock {DART}: Open-Domain Structured Data Record to Text Generation.
\newblock In \emph{Proc. of NAACL}.

\bibitem[{Pasupat and Liang(2015)}]{wikiTQ}
Pasupat, P.; and Liang, P. 2015.
\newblock Compositional Semantic Parsing on Semi-Structured Tables.
\newblock In \emph{Proc. of ACL}.

\bibitem[{Qi et~al.(2022)Qi, Tang, He, Wan, Cheng, Zhou, Wang, Zhang, and
  Lin}]{rasat}
Qi, J.; Tang, J.; He, Z.; Wan, X.; Cheng, Y.; Zhou, C.; Wang, X.; Zhang, Q.;
  and Lin, Z. 2022.
\newblock RASAT: Integrating Relational Structures into Pretrained Seq2Seq
  Model for Text-to-SQL.
\newblock In \emph{Proc. of EMNLP}.

\bibitem[{Qin et~al.(2022{\natexlab{a}})Qin, Hui, Wang, Yang, Li, Li, Geng,
  Cao, Sun, Si, Huang, and Li}]{t2s_survey}
Qin, B.; Hui, B.; Wang, L.; Yang, M.; Li, J.; Li, B.; Geng, R.; Cao, R.; Sun,
  J.; Si, L.; Huang, F.; and Li, Y. 2022{\natexlab{a}}.
\newblock A Survey on Text-to-SQL Parsing: Concepts, Methods, and Future
  Directions.
\newblock In \emph{arXiv:2208.13629}.

\bibitem[{Qin et~al.(2022{\natexlab{b}})Qin, Wang, Hui, Geng, Cao, Yang, Sun,
  and Li}]{sdcup}
Qin, B.; Wang, L.; Hui, B.; Geng, R.; Cao, Z.; Yang, M.; Sun, J.; and Li, Y.
  2022{\natexlab{b}}.
\newblock Linking-Enhanced Pre-Training for Table Semantic Parsing.
\newblock In \emph{arXiv:2111.09486}.

\bibitem[{Qin et~al.(2022{\natexlab{c}})Qin, Wang, Hui, Li, Wei, Li, Huang, Si,
  Yang, and Li}]{sun}
Qin, B.; Wang, L.; Hui, B.; Li, B.; Wei, X.; Li, B.; Huang, F.; Si, L.; Yang,
  M.; and Li, Y. 2022{\natexlab{c}}.
\newblock SUN: Exploring Intrinsic Uncertainties in Text-to-SQL Parsers.
\newblock In \emph{Proc. of COLING}.

\bibitem[{Raffel et~al.(2020)Raffel, Shazeer, Roberts, Lee, Narang, Matena,
  Zhou, Li, and Liu}]{t5-jmir-2020}
Raffel, C.; Shazeer, N.; Roberts, A.; Lee, K.; Narang, S.; Matena, M.; Zhou,
  Y.; Li, W.; and Liu, P.~J. 2020.
\newblock Exploring the Limits of Transfer Learning with a Unified Text-to-Text
  Transformer.
\newblock \emph{Journal of Machine Learning Research}.

\bibitem[{Rubin and Berant(2021)}]{smbop}
Rubin, O.; and Berant, J. 2021.
\newblock {S}m{B}o{P}: Semi-autoregressive Bottom-up Semantic Parsing.
\newblock In \emph{Proc. of NAACL}.

\bibitem[{Scholak, Schucher, and Bahdanau(2021)}]{PICARD}
Scholak, T.; Schucher, N.; and Bahdanau, D. 2021.
\newblock {PICARD}: Parsing Incrementally for Constrained Auto-Regressive
  Decoding from Language Models.
\newblock In \emph{Proc. of EMNLP}.

\bibitem[{Shaw et~al.(2021)Shaw, Chang, Pasupat, and
  Toutanova}]{shaw-etal-2021-compositional}
Shaw, P.; Chang, M.-W.; Pasupat, P.; and Toutanova, K. 2021.
\newblock Compositional Generalization and Natural Language Variation: Can a
  Semantic Parsing Approach Handle Both?
\newblock In \emph{Proc. of ACL}.

\bibitem[{Shazeer and Stern(2018)}]{adafactor}
Shazeer, N.; and Stern, M. 2018.
\newblock Adafactor: Adaptive Learning Rates with Sublinear Memory Cost.
\newblock In \emph{Proc. of ICML}.

\bibitem[{Talmor and Berant(2018)}]{kgqa}
Talmor, A.; and Berant, J. 2018.
\newblock The Web as a Knowledge-Base for Answering Complex Questions.
\newblock In \emph{Proc. of NAACL}.

\bibitem[{Vaswani et~al.(2017)Vaswani, Shazeer, Parmar, Uszkoreit, Jones,
  Gomez, Kaiser, and Polosukhin}]{transformer}
Vaswani, A.; Shazeer, N.; Parmar, N.; Uszkoreit, J.; Jones, L.; Gomez, A.~N.;
  Kaiser, L.; and Polosukhin, I. 2017.
\newblock Attention is All you Need.
\newblock In \emph{Proc. of NeurIPS}.

\bibitem[{Velickovic et~al.(2018)Velickovic, Cucurull, Casanova, Romero,
  Li{\`{o}}, and Bengio}]{gat}
Velickovic, P.; Cucurull, G.; Casanova, A.; Romero, A.; Li{\`{o}}, P.; and
  Bengio, Y. 2018.
\newblock Graph Attention Networks.
\newblock In \emph{Proc. of ICLR}.

\bibitem[{Wang et~al.(2020{\natexlab{a}})Wang, Shin, Liu, Polozov, and
  Richardson}]{wang-etal-2020-rat}
Wang, B.; Shin, R.; Liu, X.; Polozov, O.; and Richardson, M.
  2020{\natexlab{a}}.
\newblock {RAT-SQL}: Relation-Aware Schema Encoding and Linking for
  Text-to-{SQL} Parsers.
\newblock In \emph{Proc. of ACL}.

\bibitem[{Wang et~al.(2020{\natexlab{b}})Wang, Shen, Yang, Quan, and
  Wang}]{rgat}
Wang, K.; Shen, W.; Yang, Y.; Quan, X.; and Wang, R. 2020{\natexlab{b}}.
\newblock Relational Graph Attention Network for Aspect-based Sentiment
  Analysis.
\newblock In \emph{Proc. of ACL}.

\bibitem[{Wang et~al.(2022)Wang, Qin, Hui, Li, Yang, Wang, Li, Huang, Si, and
  Li}]{proton}
Wang, L.; Qin, B.; Hui, B.; Li, B.; Yang, M.; Wang, B.; Li, B.; Huang, F.; Si,
  L.; and Li, Y. 2022.
\newblock Proton: Probing Schema Linking Information from Pre-trained Language
  Models for Text-to-SQL Parsing.
\newblock In \emph{Proc. of KDD}.

\bibitem[{Wolf et~al.(2020)Wolf, Debut, Sanh, Chaumond, Delangue, Moi, Cistac,
  Rault, Louf, Funtowicz, Davison, Shleifer, von Platen, Ma, Jernite, Plu, Xu,
  Le~Scao, Gugger, Drame, Lhoest, and Rush}]{hf-transformers}
Wolf, T.; Debut, L.; Sanh, V.; Chaumond, J.; Delangue, C.; Moi, A.; Cistac, P.;
  Rault, T.; Louf, R.; Funtowicz, M.; Davison, J.; Shleifer, S.; von Platen,
  P.; Ma, C.; Jernite, Y.; Plu, J.; Xu, C.; Le~Scao, T.; Gugger, S.; Drame, M.;
  Lhoest, Q.; and Rush, A. 2020.
\newblock Transformers: State-of-the-Art Natural Language Processing.
\newblock In \emph{Proc. of EMNLP}.

\bibitem[{Xie et~al.(2022)Xie, Wu, Shi, Zhong, Scholak, Yasunaga, Wu, Zhong,
  Yin, Wang, Zhong, Wang, Li, Boyle, Ni, Yao, Radev, Xiong, Kong, Zhang, Smith,
  Zettlemoyer, and Yu}]{UnifiedSKG}
Xie, T.; Wu, C.~H.; Shi, P.; Zhong, R.; Scholak, T.; Yasunaga, M.; Wu, C.-S.;
  Zhong, M.; Yin, P.; Wang, S.~I.; Zhong, V.; Wang, B.; Li, C.; Boyle, C.; Ni,
  A.; Yao, Z.; Radev, D.; Xiong, C.; Kong, L.; Zhang, R.; Smith, N.~A.;
  Zettlemoyer, L.; and Yu, T. 2022.
\newblock UnifiedSKG: Unifying and Multi-Tasking Structured Knowledge Grounding
  with Text-to-Text Language Models.
\newblock \emph{ArXiv preprint}.

\bibitem[{Xu, Liu, and Song(2017)}]{xu2017sqlnet}
Xu, X.; Liu, C.; and Song, D. 2017.
\newblock Sqlnet: Generating structured queries from natural language without
  reinforcement learning.
\newblock \emph{ArXiv preprint}.

\bibitem[{Yaghmazadeh et~al.(2017)Yaghmazadeh, Wang, Dillig, and
  Dillig}]{yaghmazadeh2017sqlizer}
Yaghmazadeh, N.; Wang, Y.; Dillig, I.; and Dillig, T. 2017.
\newblock SQLizer: query synthesis from natural language.
\newblock \emph{Proceedings of the ACM on Programming Languages}.

\bibitem[{Ye et~al.(2022)Ye, Zhang, Deng, Chen, Chen, Xiong, Chen, and
  Chen}]{ontology}
Ye, H.; Zhang, N.; Deng, S.; Chen, X.; Chen, H.; Xiong, F.; Chen, X.; and Chen,
  H. 2022.
\newblock Ontology-enhanced Prompt-tuning for Few-shot Learning.
\newblock In \emph{WWW '22: The ACM Web Conference 2022, Virtual Event, Lyon,
  France, April 25 - 29, 2022}.

\bibitem[{Yu et~al.(2018{\natexlab{a}})Yu, Li, Zhang, Zhang, and
  Radev}]{yu-etal-2018-typesql}
Yu, T.; Li, Z.; Zhang, Z.; Zhang, R.; and Radev, D. 2018{\natexlab{a}}.
\newblock {T}ype{SQL}: Knowledge-Based Type-Aware Neural Text-to-{SQL}
  Generation.
\newblock In \emph{Proc. of NAACL}.

\bibitem[{Yu et~al.(2021)Yu, Zhang, Polozov, Meek, and Awadallah}]{score}
Yu, T.; Zhang, R.; Polozov, A.; Meek, C.; and Awadallah, A., Hassan. 2021.
\newblock SCoRe: Pre-Training for Context Representation in Conversational
  Semantic Parsing.
\newblock In \emph{Proc. of ICLR}.

\bibitem[{Yu et~al.(2018{\natexlab{b}})Yu, Zhang, Yang, Yasunaga, Wang, Li, Ma,
  Li, Yao, Roman, Zhang, and Radev}]{yu-etal-2018-spider}
Yu, T.; Zhang, R.; Yang, K.; Yasunaga, M.; Wang, D.; Li, Z.; Ma, J.; Li, I.;
  Yao, Q.; Roman, S.; Zhang, Z.; and Radev, D. 2018{\natexlab{b}}.
\newblock {S}pider: A Large-Scale Human-Labeled Dataset for Complex and
  Cross-Domain Semantic Parsing and Text-to-{SQL} Task.
\newblock In \emph{Proc. of EMNLP}.

\bibitem[{Zelle and Mooney(1996)}]{geoquery}
Zelle, J.~M.; and Mooney, R.~J. 1996.
\newblock Learning to Parse Database Queries Using Inductive Logic Programming.
\newblock In \emph{Proc. of AAAI}.

\bibitem[{Zhong et~al.(2020)Zhong, Lewis, Wang, and Zettlemoyer}]{GAZP}
Zhong, V.; Lewis, M.; Wang, S.~I.; and Zettlemoyer, L. 2020.
\newblock Grounded Adaptation for Zero-shot Executable Semantic Parsing.
\newblock In \emph{Proc. of EMNLP}.

\bibitem[{Zhong, Xiong, and Socher(2017)}]{wikiSQL}
Zhong, V.; Xiong, C.; and Socher, R. 2017.
\newblock Seq{2SQL}: Generating Structured Queries from Natural Language using
  Reinforcement Learning.
\newblock In \emph{CoRR abs/1709.00103}.

\end{thebibliography}

\newpage

\newpage
\appendix

\begin{table*}[t]
\centering
\resizebox{0.95\hsize}{!}{
\begin{tabular}{llll}
\toprule
Source $x$ & Target $y$ & Relation Type & Description \\
\midrule
Question & Question & \textsc{Modifier} & y is a modifier of x. \\
Question & Question & \textsc{Argument} & y is the source token of x under the syntax dependency outside of modifier. \\  
Question & Question & \textsc{Distance-1} & y is the nearest (1-hop) neighbor of x. \\ 

\midrule
Column & Column & \textsc{Foreign-Key} & y is the foreign key of x. \\
Column & Column & \textsc{Same-Table} & x and y appears in the same table. \\
Column & * & \textsc{Bridge} & x and y are linked when y is the special column token `*'. \\
\midrule
Table & Column & \textsc{Has} & The column y belongs to the table x. \\
Table & Column & \textsc{Primary-Key} & The column y is the primary key of the table x. \\
Table & * & \textsc{Bridge} & x and y are connected when y is the special column token `*'. \\
\midrule
Question & Table & \textsc{Exact-Match} & x is part of y, and y is a span of the entire question. \\
Question & Table & \textsc{Partial-Match} & x is part of y, but the entire question does not contain y. \\
\midrule
Question & Column & \textsc{Exact-Match} & x is part of y, and y is a span of the entire question. \\
Question & Column & \textsc{Partial-Match} & x is part of y, but the entire question does not contain y. \\
Question & Column & \textsc{Value-Match} & x is part of the candidate cell values of column y. \\
Question & * & \textsc{Bridge} & x and y are linked when y is the special column token `*'. \\
 \bottomrule
\end{tabular}
}
\caption{The checklist of main types of relations used in \graphix-T5. All relations above are asymmetric.}
\label{relation}
\end{table*}

\section{Fine-grained Syntax Relations}
The previous work \citep{lgesql, wang-etal-2020-rat}, which employed distances as the only relationship between tokens when constructing a graph, was unable to account for the deterministic relationships between tokens. For example, Given two sentences with the same meanings: \texttt{"List names of students who are not from France."};\texttt{ "What are the names of students whose nationality is not France?"}. The relation between \texttt{not} and \texttt{France} should be the same in these two sentences. However, it is represented as two different relations: \textsc{Distance-2} and \textsc{Distance-1} respectively according to their defined relations, which will lead PLMs to learn the wrong relation representations. Even though \citet{s2sql} proposed Forward and Backward as the additional abstract correlations of question tokens, it is still hard to discern the more important relations that is useful to text-to-SQL. In this work, we observe that that nouns and other tokens that potentially indicate characteristics of the nouns can help the model to find the corresponding database items. In order to achieve this goal, we cluster dependency parsing relations manually into two new categories of syntax relations: \textsc{\textbf{Modifier}} and \textsc{\textbf{Argment}}. As shown in Table \ref{relation}, the \textsc{\textbf{Modifier}} denotes that some properties of the source token node are being modified by the target token node. For example, in the phrase \texttt{Female Students}, the \texttt{Female} is a modifier of the \texttt{Students}; the \texttt{Production} is the modifier of the token \texttt{Time} in the phrase \texttt{Production Time}. All other dependency parsing relations will be marked as \textsc{\textbf{Argment}}.

\section{Leadboard Result}
After being equipped with PICARD, \graphix-T5 achieves the No.1 on \spider testing leaderboard with the clear margin, as shown in the table \ref{tab:test_em} and table \ref{tab:test_ex}.
\begin{table}[h]
\centering 
    \begin{tabular}{l|cc}
    \toprule
    \textbf{Model} & \textbf{Dev} & \textbf{Test} \\
    \midrule\midrule
    \multicolumn{3}{c}{\textbf{w/ Encoder-based PLM}} \\
    \midrule
    RATSQL + BERT & 69.7 & 65.6 \\
    RATSQL + GRAPPA & 73.4  & 69.6 \\
    GAZP + BERT & 59.1 & 53.3 \\
    BRIDGE v2 + BERT & 70.0  & 65.0  \\
    NatSQL + GAP & 73.7  & 68.7 \\
    SMBOP + GRAPPA & 74.7 & 69.7 \\
    LGESQL + ELECTRA & 75.1 & 72.0 \\
    S$^{2}$SQL + ELECTRA & 76.4 & 72.1 \\
    \midrule\midrule
    \multicolumn{3}{c}{\textbf{w/ Text-to-Text PLM: \textsc{T5}}} \\
    \midrule
    PICARD + T5-3B & 75.5  & 71.9  \\
    UnifiedSKG + T5-3B & 71.8 & - \\
    RASAT + PICARD + T5-3B & 75.3  & 70.9 \\
    \midrule
    \textbf{\graphix + PICARD + T5-3B} & \textbf{77.1} & \textbf{74.0} \\
    \bottomrule
    \end{tabular}%
    \caption{Results of Exact Match (EM) on \textsc{Spider} test.}
  \label{tab:test_em}
\end{table}%

\begin{table}[h]
\centering 
    \begin{tabular}{l|cc}
    \toprule
    \textbf{Model} & \textbf{Dev} & \textbf{Test} \\
    \midrule\midrule
    \multicolumn{3}{c}{\textbf{w/ Encoder-based PLM}} \\
    \midrule
    GAZP + BERT & 59.2 & 53.5 \\
    BRIDGE v2 + BERT & 68.3 & 64.3  \\
    NatSQL + GAP & 75.0  & 73.3 \\
    SMBOP + GRAPPA & 75.0 & 71.1 \\
    \midrule\midrule
    \multicolumn{3}{c}{\textbf{w/ Text-to-Text PLM: \textsc{T5}}} \\
    \midrule
    PICARD + T5-3B & 79.3  & 75.1  \\
    UnifiedSKG + T5-3B & 74.4 & - \\
    RASAT + PICARD + T5-3B & 80.5 & 75.5 \\
    \midrule
    \textbf{\graphix + PICARD + T5-3B} & \textbf{81.0} & \textbf{77.6} \\
    \bottomrule
    \end{tabular}%
    \caption{Results of Execution (EX) on \textsc{Spider} test.}
  \label{tab:test_ex}
\end{table}%

\end{document}